\documentclass{article} 
\usepackage{iclr2020_conference,times}

\iclrfinalcopy

\usepackage{amsmath,amsfonts,bm}

\def\eqref#1{equation~\ref{#1}}

\def\1{\bm{1}}

\DeclareMathAlphabet{\mathsfit}{\encodingdefault}{\sfdefault}{m}{sl}
\SetMathAlphabet{\mathsfit}{bold}{\encodingdefault}{\sfdefault}{bx}{n}

\usepackage{hyperref}
\usepackage{url}
\usepackage{graphicx}

\title{
Hyper-spectral NIR and MIR Data and Optimal Wavebands for Detection of Apple Tree Diseases}

\author{\textsuperscript{1}Dmitrii Shadrin\thanks{corresponding author}, \textsuperscript{1}Mariia Pukalchik, \textsuperscript{2,3}Anastasia Uryasheva, \textsuperscript{2}Evgeny Tsykunov,\\
\textbf{\textsuperscript{2}Grigoriy Yashin,} \textbf{\textsuperscript{3}{Nikita Rodichenko},}
\textbf{\textsuperscript{2}Dzmitry Tsetserukou}\\
\textsuperscript{1}Center for Computational and Data-Intensive Science and Engineering, \textsuperscript{2}Space Center \\
Skolkovo Institute of Science and Technology (Skoltech), Moscow, Russia\\
\textsuperscript{3}Tsuru Robotics (tsapk llc.), Moscow, Russia\\
\texttt{dmitry.shadrin@skolkovotech.ru} \\
}

\begin{document}

\maketitle

\begin{abstract}

Plant diseases can lead to dramatic losses in yield and quality of food, becoming a problem of high priority for farmers. Apple scab, moniliasis, and powdery mildew are the most significant apple tree diseases worldwide and may cause between 50\% and 60\% in yield losses annually; they are controlled by fungicide use with huge financial and time expenses. This research proposes a modern approach for analyzing the spectral data in Near-Infrared and Mid-Infrared ranges of the apple tree diseases at different stages. Using the obtained spectra, we found optimal spectral bands for detecting particular disease and discriminating it from other diseases and healthy trees. The proposed instrument will provide farmers with accurate, real-time information on different stages of apple tree diseases, enabling more effective timing, and selecting the fungicide application, resulting in better control and increasing yield. The obtained dataset as well as scripts in \textit{Matlab} for processing data and finding optimal spectral bands are available via the link: \url{https://yadi.sk/d/ZqfGaNlYVR3TUA}

\end{abstract}

\section{Introduction}
Automatic systems for visual monitoring, along with the newest machine learning techniques, are very powerful tools for monitoring and assessing the quality and quantity of agricultural production. At the same time, these systems could help farmers to save their yield losses due to remote monitoring. The possible application of plant growth optimization with advanced non-invasive technologies for disease detection has previously described in \cite{hanan2017greenhouses}, \cite{park2011study}, \cite{mahlein2012recent} and \cite{rumpf2010early}. Nowadays, farmers are going to improve the overall quality and quantity of their harvest, to predict maturation rate, and to check the productivity of plantings together with decreasing the workload among food production \citep{fan2011improving}. Harvest losses due to pests and diseases are a major threat and costing billions of dollars every year  \citep{rubatzky2012world}. Meanwhile, most of the current approaches for monitoring and detecting diseases performed by human hands, which result in approximate assessment and special tools metrics for evaluation, are rarely used as it is time-consuming and it requires lots of human resources. In addition, workers without special equipment couldn't cover and analyze all plantations or greenhouses due to their large sizes. This problem leads to a loss of important information about the epicenters of new plants diseases and its real diversity. In addition, this case makes it challenging to build a comprehensive map of diseases that occur in space and time. Diseases in a plant can occur quickly in real field environment: e.g., common fungal diseases such as apple scab can attack plant in two weeks \citep{vanderplank2012disease}, thus, only real-time monitoring can help to detect diseases at early stages and to optimize fungicide application \citep{sophie2010disease}.

Computer vision systems go beyond human capabilities and help to evaluate long-term processes and events occurring in the whole electromagnetic spectrum. Among them, hyper-spectral systems provide features that can be used as fingerprints of plant diseases at a certain wavelength. This finding can be used as a tool for developing new computer vision systems adapted for specific agriculture purposes. It is also important to discriminate one disease from another one by using classification approaches jointly with image processing \citep{khan2019optimized}, \citep{pantazi2019automated}, \citep{sarfraz2014computer}, \citep{patil2011advances}. Nowadays, deep learning models for plant disease detection, that include long-short term memory neural networks (LSTM) coupled with convolution neural networks (CNN), in particular for the detection of apple scab, are showing good results \cite{baranwal2019deep}, \citet{turkoglu2019multi} and \cite{ferentinos2018deep}. Neural networks have also showed their usefulness for plant disease detection basing on the hyper-spectral data \cite{golhani2018review}. There are several recently published datasets that allow training deep learning models, such as \cite{parraga2019rocole},  \cite{nouri2018near}. However, in the available near-infrared hyperspectral dataset \cite{nouri2018near}, there are not so many obtained data and spectra for training machine learning (ML) algorithms, and also, waveband is narrow. Using the hyper-spectral data, we can improve the accuracy if optimal spectra are known. Also, it is not necessary to collect a huge dataset, as features have already been extracted. Thus, it will allow using simpler algorithms for detecting plant diseases, which in turn will allow running models on the low power embedded devices. In this work, we use spectral reflective properties of three apple tree fungal diseases (apple scab caused by Venturia inaequals, Moniliasis caused by Monilia, and powdery mildew caused by Erysiphales) to reveal optimal wavebands of this disease at different stages.

\section{Dataset Collection and Spectral Analysis}
\label{sec:Dataset_collection}

\subsection{Spectral Data on Apple Scab}

\begin{figure}[h]
\begin{centering}
\begin{tabular}{cc}
\includegraphics[width=0.45\linewidth]{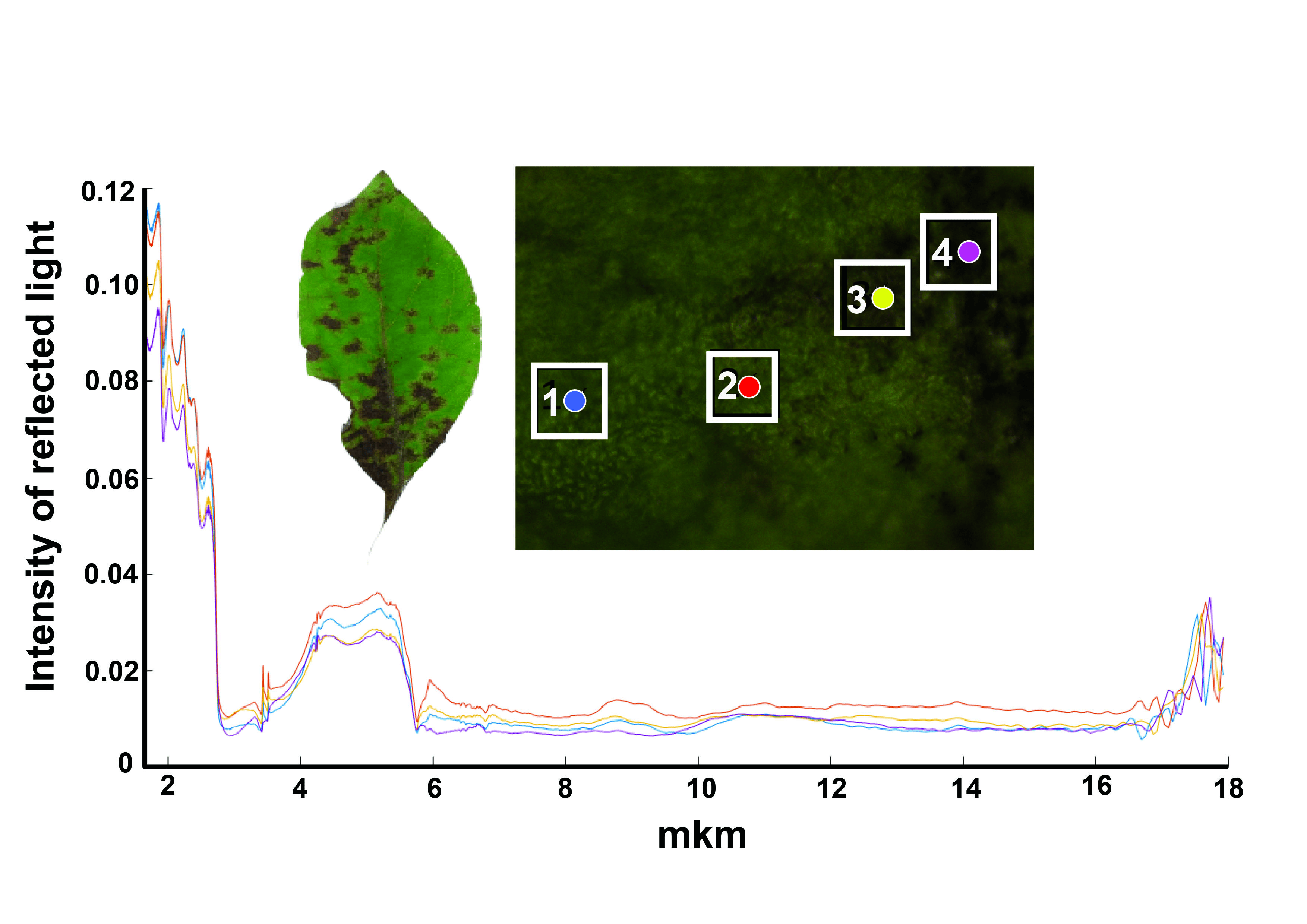}&
\includegraphics[width=0.45\linewidth]{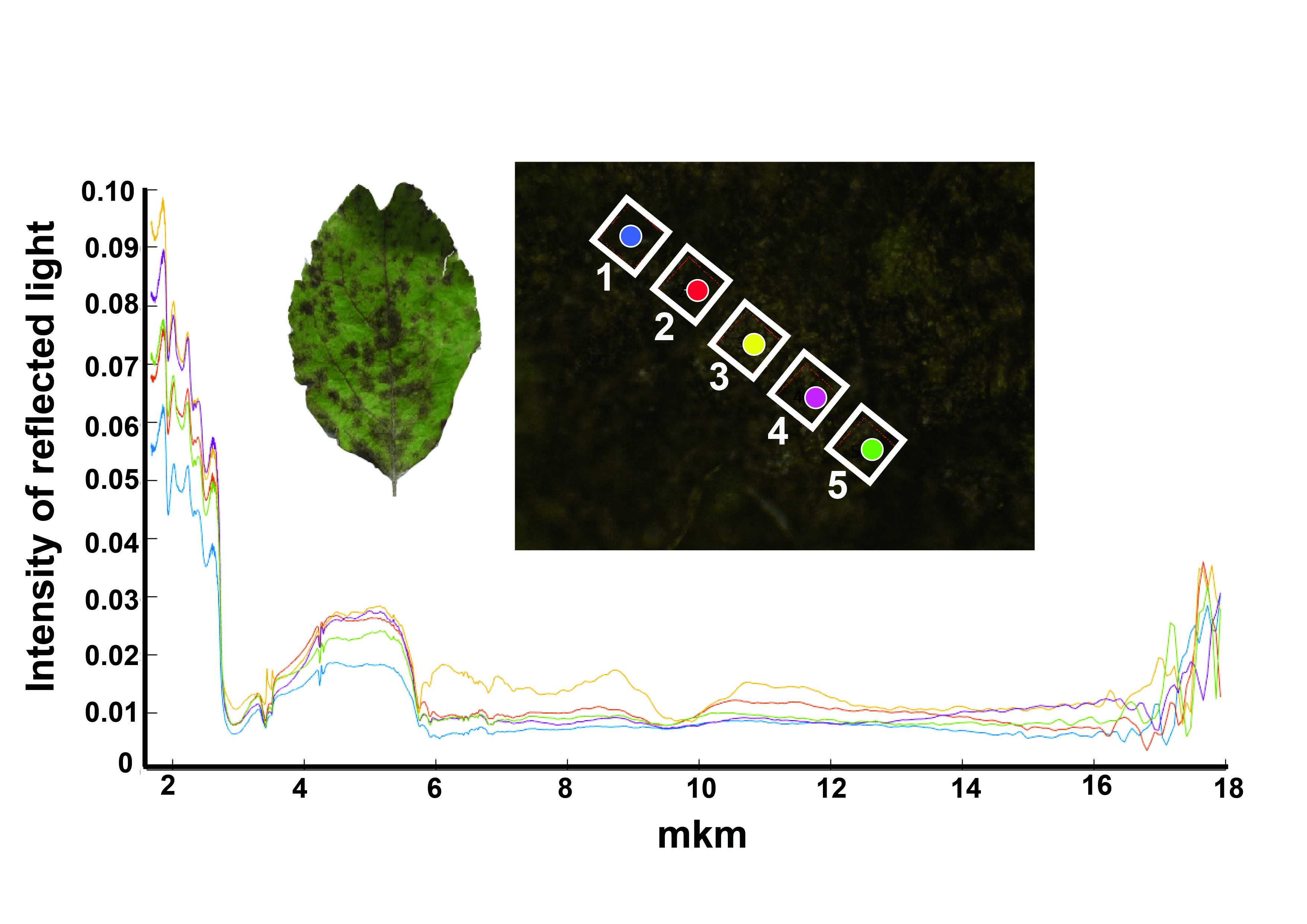}\\
(a)&(b) \\
\includegraphics[width=0.45\linewidth]{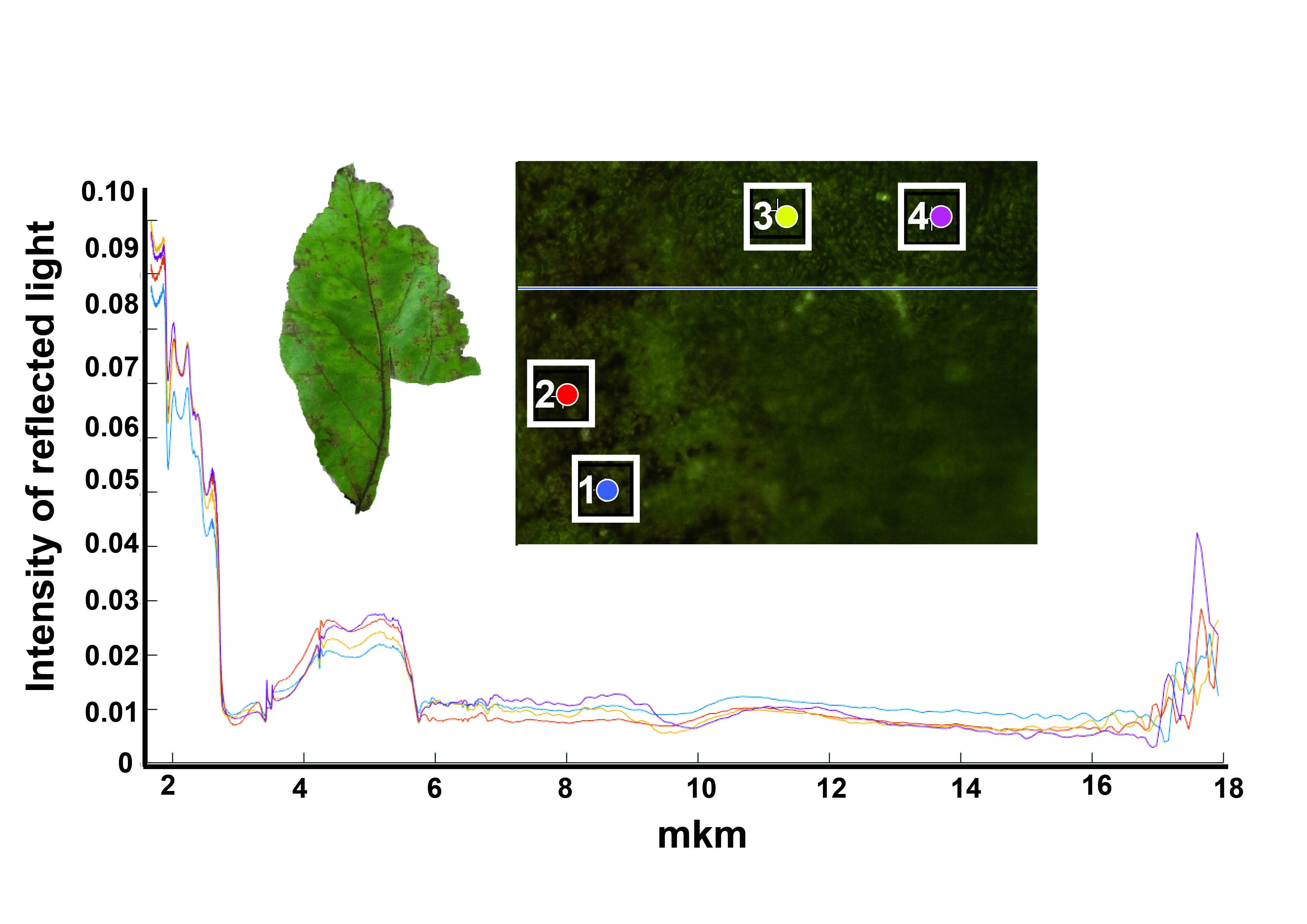}&
\includegraphics[width=0.45\linewidth]{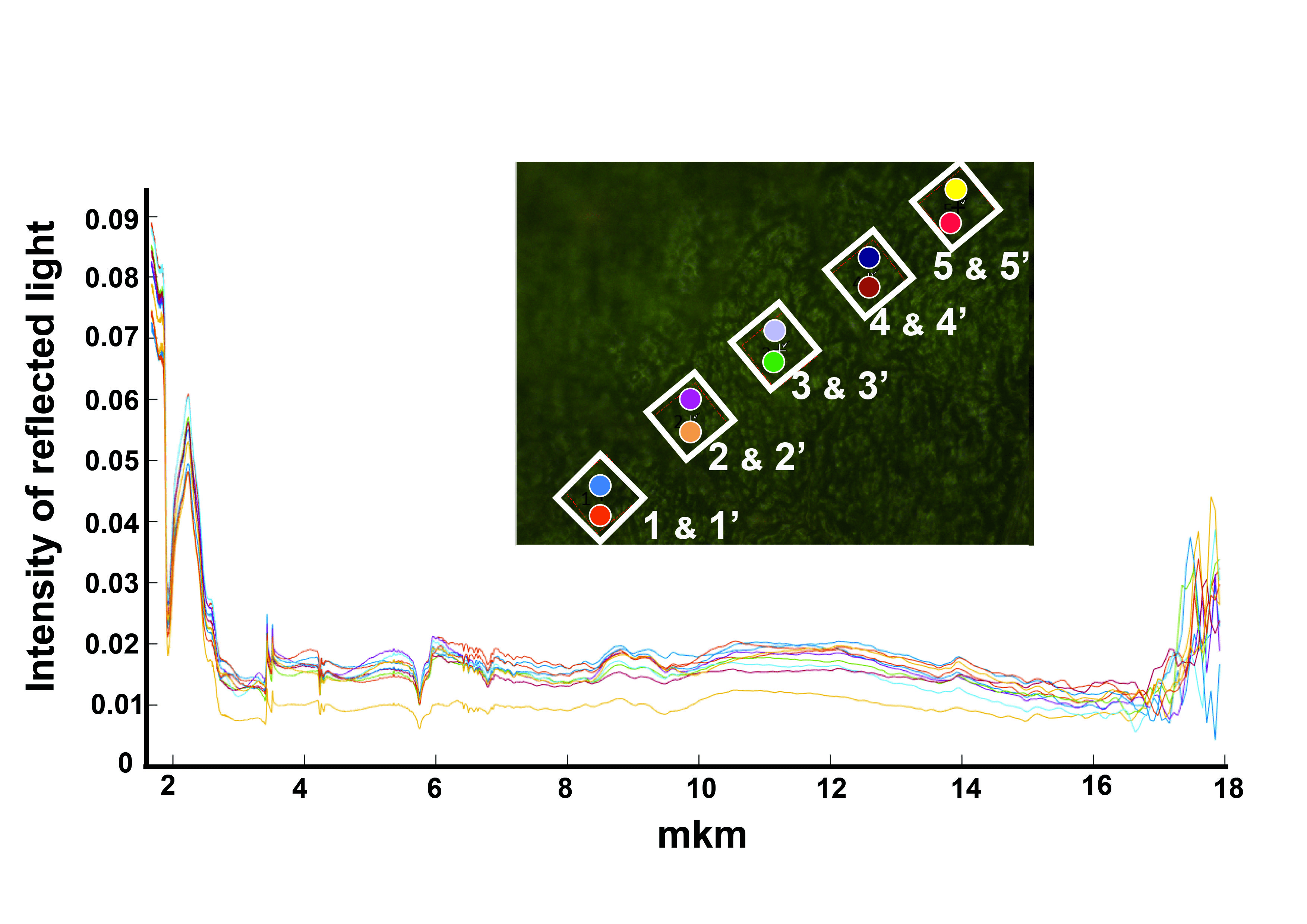}\\
(c)&(d) \\
\end{tabular}
\caption{Obtained spectra and leaf samples for: (a), (b) four sub-regions of infected leaves respectively, (c) four sub-regions of cured leaf, (d) 8 sub-regions of healthy leaf.}
\label{fig:Raw_Spectra}
\end{centering}
\end{figure}

Apple scab is a disease of apple trees caused by fungi and it affects leaves and fruits. Seven samples of the apple leaves were selected for obtaining spectra: four leaves were infected with apple scab at different stages, two leaves were cured of apple scab, and one healthy leaf was used as a reference. On each leaf, in a small region of ~1000 $\mu m^2$, 5-6 sub-regions of ~10 $\mu m^2$ were allocated. For each of these sub-regions, the spectrum of reflected light was measured in the infrared region of 1.6-18 $\mu m$. The achieved spectral data were used to distinguish between infected, diseased, cured, and healthy leaves. The 35 spectra were obtained to get reliable results. The examples of leaf spectra and leaf samples are presented in the Fig. \ref{fig:Raw_Spectra}. Importantly, we noticed that there is no much difference between the recorded spectra of two regions of one infected leaf: regions with visually observed signs of scab, and the other two on which there are no visible signs of the scab, i.e., regions at earliest stage of disease (Fig. \ref{fig:Raw_Spectra}). Similar results were obtained for fully damaged leaf by apple scab (Fig. \ref{fig:Raw_Spectra}b) and for treated scab leaf (Fig. \ref{fig:Raw_Spectra}c). 
Unsurprisingly, spectra for healthy leaf significantly differentiate from infected samples in all investigated sub-regions (Fig. \ref{fig:Raw_Spectra}d). This observation opens a gate at remote detection of apple scab. Using the same approach, we also collected data on leaves infected by moniliasis and powdery mildew. In total, we obtained 20 spectra for moniliasis and 16 spectra for powdery mildew. Note, that we release the full dataset, which may provide the chance to further expand on our study.

\label{sec:Apple_scab_laboratory}

\section{Finding Optimal Spectral Wavebands for Apple Tree Disease Detection}
\label{sec:optimal}

\begin{figure}[h]
\begin{center}
\includegraphics[width=0.9\linewidth]{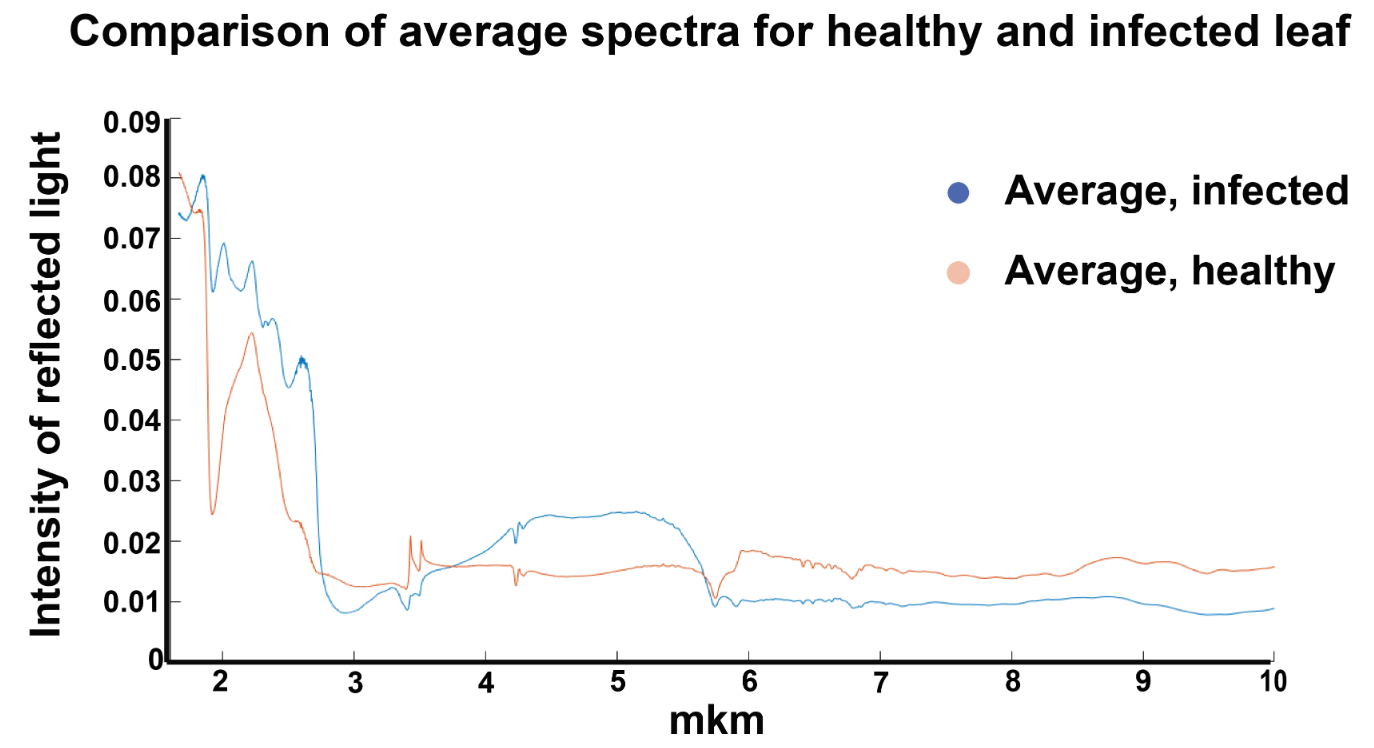}
\end{center}
\caption{Averaged spectra for infected and healthy leaves.}
\label{fig:averaged_spectra}
\end{figure}

\begin{figure}[htbp]
\begin{center}
\includegraphics[width=0.9\linewidth]{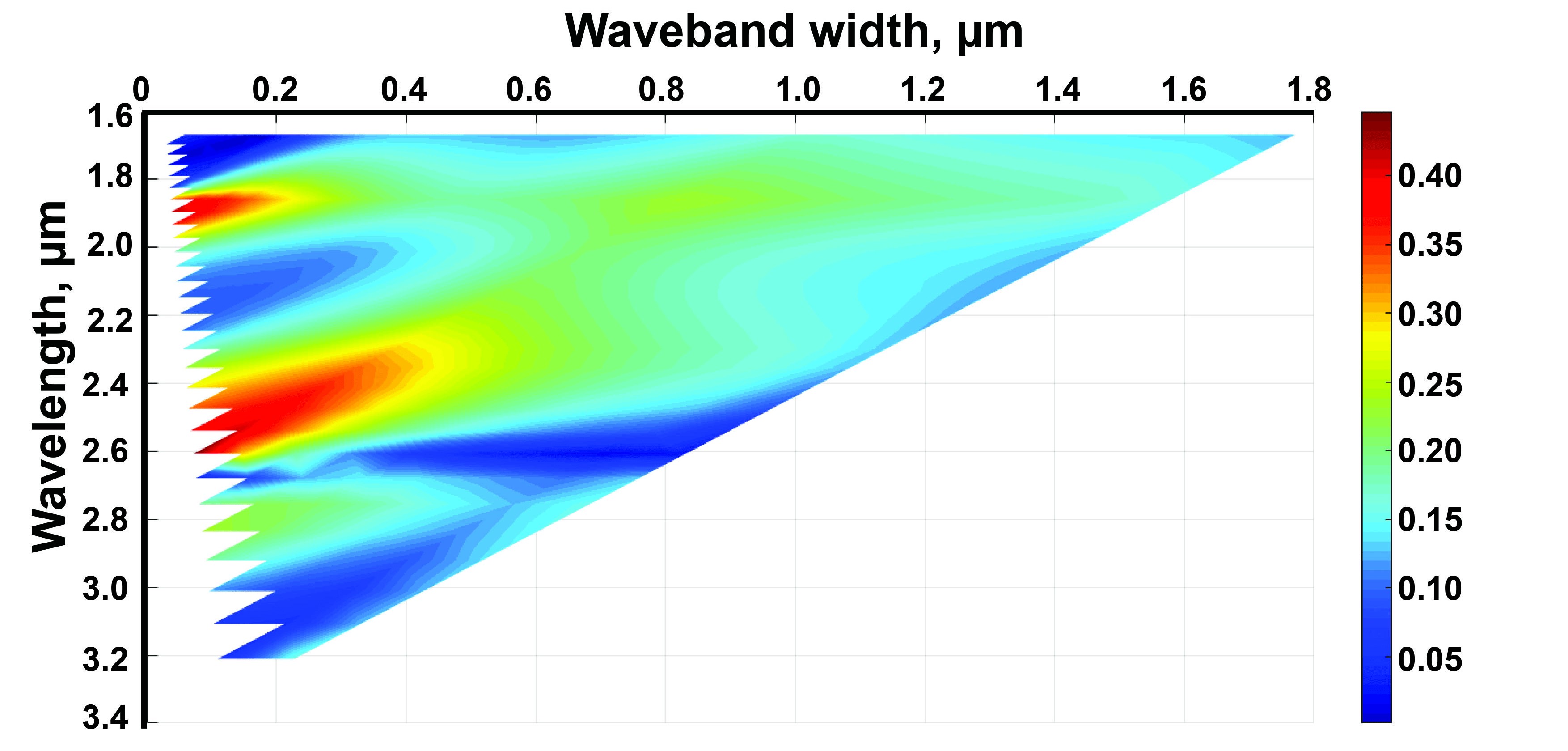}
\end{center}
\caption{Distribution of the discriminative coefficient for different spectral wavebands, y-axis is wavelength from which waveband started, x-axis is the width of waveband.}
\label{fig:optimal_wavebands}
\end{figure}

The averaged spectra for healthy and diseased leaves are shown in the Fig. \ref{fig:averaged_spectra}. It used to reveal the optimal bandwidth for suitable infrared cameras selection for in vivo studies.
Based on the averaged spectra presented in Fig. \ref{fig:averaged_spectra}, discriminative coefficients were simulated in order to solve the classification problem. Using them, optimal bandwidths for disease detection were defined. The simulation was carried out in \textit{MATLAB} development environment. It was proposed to introduce a new discriminative coefficient similar in structure to the normalized difference vegetation index (NDVI). Our proposed coefficient is the absolute difference of the reflection of the bands divided by their sum for normalization 
(\ref{eq:disc_coef}) (area between two spectra divided by the sum of the areas under each spectrum). 
\begin{equation} \label{eq:disc_coef}
Discriminative Coef(i,j) = \frac{|AUC_1(i,j) - AUC_2(i,j)|}{AUC_1(i,j) + AUC_2(i,j)}
\end{equation}
where $AUC$ is the area under an averaged spectrum of waveband, $i$ is the wavelength from each waveband started, and $j$ is the width of waveband.
The coefficients were calculated for the available spectra for all wavelengths and for all possible bands. The minimum bandwidth step used in the simulation is 50 $nm$, due to the fact that more narrow-band cameras do not seem possible to use in the field. The results are presented in the form of 2-d graph (see Fig. \ref{fig:optimal_wavebands}), representing the value of the discrimination coefficient for apple scab. It should be noticed that in the Fig. \ref{fig:optimal_wavebands} simulation presented only up to 3.2 $\mu m$, because the value of discriminative coefficient for larger wavelength is not significant for this particular case (see Fig. \ref{fig:averaged_spectra}). Areas in Fig. \ref{fig:optimal_wavebands} with relatively high values of coefficient represent the most selective wavebands. It can be noticed from Fig. \ref{fig:optimal_wavebands}, that regions starting from 1.8-2.0 $\mu m$ wavelength with bandwidth 0.2-0.4 $\mu m$, also regions starting from 2.4-2.6 $\mu m$ with bandwidth 0.1-0.4 $\mu m$ have a good selective ability, since the coefficient value is relatively high. This result is very well explained by theoretical assumptions: Fungi, while destroying cells, causes decreasing of water content, thus. decreasing water absorption, which can be clearly seen in the infrared waveband of spectral region. Resulting classification bands coincide with the water absorption spectrum. The distribution of the discriminative coefficient for different wavelengths and bandwidth is shown in Fig. \ref{fig:optimal_wavebands}

Using the same approach discussed above and spectral data for other apple tree diseases, simulations of the distribution of discriminative coefficients were performed. Diagrams, similar to Fig. \ref{fig:optimal_wavebands} for discrimination of other types of diseases are presented in the shared dataset. Results of simulations are shown in Table \ref{table:summary_results}. Notably, we achieve state-of-the-art results, and optimal spectra for different diseases are not overlap.
\begin{table}[h]
\caption{Summary of the highest values of the discriminative coefficient representing the best wavebands in near and short infrared spectra for detecting apple tree diseases.}
\label{table:summary_results}
\begin{center}
\begin{tabular}{lcc}
\multicolumn{1}{c}{\bf Disease} & \multicolumn{1}{c}{\bf Waveband, $\mu m$}  & \multicolumn{1}{c}{\bf Coef.value}
\\ \hline \\
Apple scab, healthy/infected & 1.8-2; 2.4-2.8 & 0.5\\
Moniliasis, spores/infected apple & 2.8-3.1 & 0.7\\
Moniliasis, healthy/infected skin of apple & 1.6-1.8 & 0.25\\
Moniliasis, healthy/spores & 2.9-3.2; 5.9-9 & 0.8\\
Powdery mildew healthy/infected & 2.7-2.9 & 0.41\\
\end{tabular}
\end{center}
\end{table}
\section{Conclusions}

Motivated to detect early-stage disease for apple trees better, we set out a new dataset with spectral data in Near-infrared and Mid-infrared spectrum range which consists of 51 spectra in the spectrum range 1.6 - 18 $\mu m$ for different apple tree disease. We proposed a new combined approach to finding optimal wavebands for discriminating apple tree disease stages. Our approach was successfully tested on all obtained datasets for detecting three different diseases, and obtained results coincided with theoretical assumptions and showed its accuracy. Also, such features, designed from spectral data, can be useful for a deep learning approach, in case of implementation on embedded systems for field disease detection. Using these features will potentially allow to decrease the number of layers, making deployed networks more shallow and easier to run on embedded systems. We release the full datasets and \textit{Matlab} code in a shared repository and to provide the important data on Near-infrared and Mid-infrared spectrum range along with optimal bandwidth encouraging community to develop the systems for remote detection of apple tree diseases at the earliest stages. 

\section*{Acknowledgements} 
This work was funded, in part, by the Soyuz Snab LLC. The authors are especially grateful to Boris Afinogenov for his help in collection of spectral data of samples using equipment of the laboratory of Center for Photonics and Quantum Materials under the direction of Albert Nasibulin.

\bibliography{iclr2020_conference.bib}

\begin{thebibliography}{18}
\providecommand{\natexlab}[1]{#1}
\providecommand{\url}[1]{\texttt{#1}}
\expandafter\ifx\csname urlstyle\endcsname\relax
  \providecommand{\doi}[1]{doi: #1}\else
  \providecommand{\doi}{doi: \begingroup \urlstyle{rm}\Url}\fi

\bibitem[Baranwal et~al.(2019)Baranwal, Khandelwal, and
  Arora]{baranwal2019deep}
Saraansh Baranwal, Siddhant Khandelwal, and Anuja Arora.
\newblock Deep learning convolutional neural network for apple leaves disease
  detection.
\newblock \emph{Available at SSRN 3351641}, 2019.

\bibitem[Fan et~al.(2011)Fan, Shen, Yuan, Jiang, Chen, Davies, and
  Zhang]{fan2011improving}
Mingsheng Fan, Jianbo Shen, Lixing Yuan, Rongfeng Jiang, Xinping Chen,
  William~J Davies, and Fusuo Zhang.
\newblock Improving crop productivity and resource use efficiency to ensure
  food security and environmental quality in china.
\newblock \emph{Journal of experimental botany}, 63\penalty0 (1):\penalty0
  13--24, 2011.

\bibitem[Ferentinos(2018)]{ferentinos2018deep}
Konstantinos~P Ferentinos.
\newblock Deep learning models for plant disease detection and diagnosis.
\newblock \emph{Computers and Electronics in Agriculture}, 145:\penalty0
  311--318, 2018.

\bibitem[Golhani et~al.(2018)Golhani, Balasundram, Vadamalai, and
  Pradhan]{golhani2018review}
Kamlesh Golhani, Siva~K Balasundram, Ganesan Vadamalai, and Biswajeet Pradhan.
\newblock A review of neural networks in plant disease detection using
  hyperspectral data.
\newblock \emph{Information Processing in Agriculture}, 5\penalty0
  (3):\penalty0 354--371, 2018.

\bibitem[Hanan(2017)]{hanan2017greenhouses}
Joe~J Hanan.
\newblock \emph{Greenhouses: Advanced technology for protected horticulture}.
\newblock CRC press, 2017.

\bibitem[Khan et~al.(2019)Khan, Lali, Sharif, Javed, Aurangzeb, Haider,
  Altamrah, and Akram]{khan2019optimized}
Muhammad~Attique Khan, M~Ikram~Ullah Lali, Muhammad Sharif, Kashif Javed,
  Khursheed Aurangzeb, Syed~Irtaza Haider, Abdulaziz~Saud Altamrah, and Talha
  Akram.
\newblock An optimized method for segmentation and classification of apple
  diseases based on strong correlation and genetic algorithm based feature
  selection.
\newblock \emph{IEEE Access}, 7:\penalty0 46261--46277, 2019.

\bibitem[Mahlein et~al.(2012)Mahlein, Oerke, Steiner, and
  Dehne]{mahlein2012recent}
Anne-Katrin Mahlein, Erich-Christian Oerke, Ulrike Steiner, and Heinz-Wilhelm
  Dehne.
\newblock Recent advances in sensing plant diseases for precision crop
  protection.
\newblock \emph{European Journal of Plant Pathology}, 133\penalty0
  (1):\penalty0 197--209, 2012.

\bibitem[Nouri et~al.(2018)Nouri, Gorretta, Vaysse, Giraud, Germain, Keresztes,
  and Roger]{nouri2018near}
Maroua Nouri, Nathalie Gorretta, Pierre Vaysse, Michel Giraud, Christian
  Germain, Barna Keresztes, and Jean-Michel Roger.
\newblock Near infrared hyperspectral dataset of healthy and infected apple
  tree leaves images for the early detection of apple scab disease.
\newblock \emph{Data in brief}, 16:\penalty0 967--971, 2018.

\bibitem[Pantazi et~al.(2019)Pantazi, Moshou, and
  Tamouridou]{pantazi2019automated}
Xanthoula~Eirini Pantazi, Dimitrios Moshou, and Alexandra~A Tamouridou.
\newblock Automated leaf disease detection in different crop species through
  image features analysis and one class classifiers.
\newblock \emph{Computers and electronics in agriculture}, 156:\penalty0
  96--104, 2019.

\bibitem[Park et~al.(2011)Park, Kang, Cho, Shin, Cho, Park, and
  Yang]{park2011study}
Dae-Heon Park, Beom-Jin Kang, Kyung-Ryong Cho, Chang-Sun Shin, Sung-Eon Cho,
  Jang-Woo Park, and Won-Mo Yang.
\newblock A study on greenhouse automatic control system based on wireless
  sensor network.
\newblock \emph{Wireless Personal Communications}, 56\penalty0 (1):\penalty0
  117--130, 2011.

\bibitem[Parraga-Alava et~al.(2019)Parraga-Alava, Cusme, Loor, and
  Santander]{parraga2019rocole}
Jorge Parraga-Alava, Kevin Cusme, Ang{\'e}lica Loor, and Esneider Santander.
\newblock Rocole: A robusta coffee leaf images dataset for evaluation of
  machine learning based methods in plant diseases recognition.
\newblock \emph{Data in brief}, 25:\penalty0 104414, 2019.

\bibitem[Patil \& Kumar(2011)Patil and Kumar]{patil2011advances}
Jayamala~K Patil and Raj Kumar.
\newblock Advances in image processing for detection of plant diseases.
\newblock \emph{Journal of Advanced Bioinformatics Applications and Research},
  2\penalty0 (2):\penalty0 135--141, 2011.

\bibitem[Rubatzky \& Yamaguchi(2012)Rubatzky and Yamaguchi]{rubatzky2012world}
Vincent~E Rubatzky and Mas Yamaguchi.
\newblock \emph{World vegetables: principles, production, and nutritive
  values}.
\newblock Springer Science \& Business Media, 2012.

\bibitem[Rumpf et~al.(2010)Rumpf, Mahlein, Steiner, Oerke, Dehne, and
  Pl{\"u}mer]{rumpf2010early}
T~Rumpf, A-K Mahlein, U~Steiner, E-C Oerke, H-W Dehne, and L~Pl{\"u}mer.
\newblock Early detection and classification of plant diseases with support
  vector machines based on hyperspectral reflectance.
\newblock \emph{Computers and Electronics in Agriculture}, 74\penalty0
  (1):\penalty0 91--99, 2010.

\bibitem[Sarfraz(2014)]{sarfraz2014computer}
Muhammad Sarfraz.
\newblock \emph{Computer Vision and Image Processing in Intelligent Systems and
  Multimedia Technologies}.
\newblock IGI Global, 2014.

\bibitem[Sophie et~al.(2010)]{sophie2010disease}
Anne Sophie et~al.
\newblock Disease decision support systems: their impact on disease management
  and durability of fungicide effectiveness.
\newblock In \emph{Fungicides}. InTech, 2010.

\bibitem[Turkoglu et~al.(2019)Turkoglu, Hanbay, and Sengur]{turkoglu2019multi}
Muammer Turkoglu, Davut Hanbay, and Abdulkadir Sengur.
\newblock Multi-model lstm-based convolutional neural networks for detection of
  apple diseases and pests.
\newblock \emph{Journal of Ambient Intelligence and Humanized Computing}, pp.\
  1--11, 2019.

\bibitem[Vanderplank(2012)]{vanderplank2012disease}
James~Edward Vanderplank.
\newblock \emph{Disease resistance in plants}.
\newblock Elsevier, 2012.

\end{thebibliography}
\bibliographystyle{iclr2020_conference}

\end{document}